\documentclass[letterpaper, 10 pt, conference]{ieeeconf}  

\IEEEoverridecommandlockouts                              

\usepackage{amsmath,amssymb,amsfonts,amsthm}
\usepackage{tabularx,booktabs, multirow}
\usepackage{graphicx}
\usepackage{bm}
\usepackage{algpseudocode}
\usepackage{algorithm}
\usepackage{physics}
\makeatletter
\def\endfigure{\end@float}
\def\endtable{\end@float}
\makeatother
\usepackage[caption=false]{subfig} 
\usepackage{xcolor}
\usepackage{threeparttable}

\theoremstyle{definition}
\newtheorem{assumptionx}{Assumption}

\DeclareMathOperator*{\argmax}{arg\,max}

\newcommand{\tp}{^\top}
\newcommand{\Fro}{_\mathrm{F}}
\DeclareDocumentCommand\diag{}{\opbraces{\operatorname{diag}}}

\title{\LARGE \bf
Hybrid Optimization-Generative Motion Planning\\ for Multiagent Affine Formations
}

\author{Zhonggang Li, Hongyu Zhou, and Raj Thilak Rajan
\thanks{This work is partially funded by the Sensor AI Lab, under the AI Labs program of Delft University of Technology, and by the RVO (Missie gedreven Onderzoek, Ontwikkeling en Innovatie (MOOI)) Aerial Quadcopter Units for Aquatic Flow Investigation and Nautical Data (AQUAFIND) project. The authors are with the Signal Processing Systems group, Department of Microelectronics, Faculty of EEMCS of Delft University of Technology, The Netherlands. Emails: \{z.li-22, r.t.rajan\}@tudelft.nl, and h.zhou-13@student.tudelft.nl.}}

\begin{document}

\maketitle
\thispagestyle{empty}
\pagestyle{empty}

\begin{abstract}
Formation maneuver control enables a multiagent system to cooperatively navigate through constrained environments in collective motion. While various formation controllers exist, they rely heavily on high-quality and dynamically feasible reference trajectories provided a priori, making formation motion planning (FMP) essential. To address the computational challenges for high-dimensional joint trajectories faced by conventional constrained optimization-based solutions, this paper introduces a hybrid optimization-generative framework with a focus on affine formations, which provide the necessary spatial flexibility. By formulating the planning task as a constrained Maximum-a-Posteriori estimation, we decouple multiagent objectives to leverage pre-trained single-agent diffusion priors for dynamic feasibility, obstacle avoidance, and smoothness. We then interleave the generative process with projected gradient descent (PGD) to strictly enforce affine geometric constraints with safety guarantees. Simulations demonstrate our framework generates highly smooth, collision-free plans that are readily trackable by common distributed controllers, and outperforms state-of-the-art planners in various metrics.
\end{abstract}

\section{Introduction}\label{sec: intro}
Formation maneuvering is an essential task for multi-robotic systems, aiming to navigate a swarm of robots safely through a potentially cluttered area while maintaining a desired range of geometries. This is often desired in applications such as cooperative object transport \cite{alonso2017multi}, satellite interferometry \cite{Zhao2023Satellite}, sensing and coverage \cite{abbasi2017team, saska2017documentation}, etc. 
Recently, deformable formations \cite{trinh2018bearing, zhao2018affine, cao2023similar} have received increasing attention due to the advantages of flexible geometries for complex environments, over conventional rigid body control  \cite{vu2023distance}. In particular, affine formations \cite{zhao2018affine, li2026geometry, aranda2023combined}, including translation, rotation, scaling, and shear, have become a popular expressive class. Predominant formation control literature solves stabilization and tracking of maneuvers assuming given references, i.e., the trajectories are assumed to be provided a priori and are inherently collision-free and dynamically feasible. Moreover, in common distributed control \cite{zhao2018affine}, followers rely solely on local information. Consequently, if the planned trajectories deviate from a strict affine transformation, the formation cannot be resolved locally, leading to severe tracking degradation. Hence, enforcing affine geometry is essential for such distributed controllers.

Generating a full set of high-quality reference trajectories subject to complex geometry constraints is difficult. Practical approaches typically decouple the problem hierarchically by pairing a simple global path planner (typically artificial potential field (APF) or A*) for the swarm centroid or a leader with a reactive local controller to resolve collisions and geometry \cite{alonso2016distributed, du2025motion, zhang2025obstacle, wang2023novel, saska2014motion}. For instance, \cite{du2025motion} plans for a formation that allows rotation and anisotropic scaling of the geometry, and it uses APF to traverse a feasible global path for the centroid with coarse shape planning, followed by a model predictive controller (MPC) that refines the shape geometry and resolves collision avoidance at runtime. This paradigm shifts the heavy sensing and computational burden of collision avoidance and geometric maintenance entirely onto the local controller. Consequently, the required complex online solvers can easily overwhelm the limited onboard resources, such as on lightweight or space robots. More fundamentally, this approach is bottlenecked by the quality of the reference trajectory, making it difficult to apply common tracking controllers such as \cite{zhao2018affine}, which will inevitably lead to suboptimal maneuvers or tracking failure.

An alternative paradigm to mitigate complexity is to plan within a low-dimensional space parameterized by the formation's macroscopic geometric variables \cite{alonso2017multi,liu2023novel,quan2023robust}. Since geometric coupling intrinsically binds the agents, a 2D affine transformation allows the entire swarm's configuration to be fully described by just 6 degrees of freedom (DOF). Leveraging this reduction, \cite{alonso2017multi,quan2023robust} optimize trajectories strictly within the 7-DOF 3D similarity transformation space, while \cite{liu2023novel} restricts planning to a $3$-DOF 2D rigid body transformation. However, planning exclusively within the geometric parameter space is highly susceptible to disturbances. Small perturbations in the macroscopic parameters are often significantly amplified when mapped to the individual agents, resulting in severe non-smoothness and dynamic infeasibility in the actual trajectory domain. As such, these geometric abstractions must be paired with advanced auxiliary space-time smoothers in the trajectory domain. For example, \cite{quan2023robust} develops a spatiotemporal polynomial optimization framework to guarantee smooth, dynamically feasible flight, embedded in the precomputed geometric references.

Recently, learning-based motion planning has received increasing attention, such as reinforcement learning-based FMP \cite{hadi2023adaptive}. In particular, generative models such as diffusion models \cite{ho2020denoising,nichol2021improved} have been successfully applied to robot motion planning \cite{janner2022diffuser, carvalho2023motion, christopher2024constrained}. The idea is to internalize the complex, non-convex motion constraints into learned structural priors through a large amount of data. Then, during inference, we sample from this learned prior with mild intervention from conventional optimization using gradient descent for task-specific objectives. Compared with conventional approaches, this optimization-generative method is computationally tractable, with better smoothness and reachability in cluttered settings \cite{carvalho2023motion}. However, this approach is currently limited to single-agent planning, as multiagent trajectory data is much more difficult to collect for an arbitrary number of mobile agents. As such, pioneering works for general multiagent motion planning with diffusion models \cite{shaoul2025multi} mostly focus on higher-level priority coordination based on single-agent priors. Moreover, diffusion-based motion planning for multiagent flexible formation maneuvers remains unexplored. 

In this work, we propose a diffusion-based motion planning framework specifically for multiagent affine formations. We formulate the FMP task as a constrained Maximum-a-Posteriori (MAP) estimation, which naturally translates to a hybrid constrained optimization with generative priors. By decoupling the high-dimensional multiagent objectives, we can leverage pre-trained single-agent diffusion priors to fulfill intra-agent requirements (such as smoothness and obstacle avoidance), while resolving inter-agent interactions, i.e., the formation, through projected gradient descent (PGD). Additionally, a safety distance among agents can be guaranteed by configuring the affine geometry parameters in the projection step. Our proposed approach avoids the need for complex reactive controllers or additional spatiotemporal trajectory smoothers as in conventional methods. The generated trajectories are smooth, strictly affine, and executable by common distributed tracking controllers.

The remainder of the paper is organized as follows. The general problem formulation and single-agent motion planning with diffusion is introduced in Section \ref{sec: fundamentals}. Then the proposed FMP framework with formation projection and safety guarantees is discussed in Section \ref{sec: fmpd}. Finally, we validate our algorithm and the quality of the planned trajectories in comparison with state-of-the-art methods in Section \ref{sec: sim}. We then conclude the discussion with insights into the future directions in Section \ref{sec: conclusion}. Throughout the paper, we adhere to the following notation conventions for mathematical clarity.\\

\noindent \emph{Notations.} Vectors and matrices are represented by lowercase and uppercase boldface letters, respectively, such as $\bm{a}$ and $\bm{A}$. Sets and graphs are represented using calligraphic letters, e.g., $\mathcal{A}$. Vectors of length $N$ of all ones and zeros are denoted by $\bm{1}_N$ and $\bm{0}_N$, respectively, and their matrix versions are similarly $\bm{1}_{M\times N}$ and $\bm{0}_{M\times N}$. An identity matrix of size $N$ is denoted by $\bm{I}_N$. The $\diag(\cdot)$ operator creates a diagonal matrix from a vector. Units are enclosed by square brackets, e.g., seconds $[s]$.

\section{PRELIMINARIES}\label{sec: fundamentals}
We consider an $N$-agent swarm in $D$-dimensional Euclidean space. A $\textit{trajectory}$ of an agent $i=1,\dots, N$ is denoted by $\bm{\tau}_i = \qty[\bm{s}_{i,0}\tp,\dots,\bm{s}_{i,H-1}\tp]\tp\in\mathbb{R}^{2DH}$, which is a collection of the agents' states $\bm{s}_{i,h}$  across a horizon $h=0,\dots,H-1$. The state contains the agent's position and velocity, i.e., $\bm{s}_{i,h} = \mqty[\bm{p}_{i,h}\tp & \dot{\bm{p}}_{i,h}\tp]\tp\in\mathbb{R}^{2D}$. The joint planning variable is then the collection of all agents' trajectories $\bm{\tau} = \qty[\bm{\tau}_1\tp,\dots,\bm{\tau}_N\tp]\tp\in\mathbb{R}^{2DHN}$. Given a set of task objectives $\mathcal{O}$, the formational motion planning (FMP) could be formulated as a constrained Maximum-a-Posteriori (MAP) problem
\begin{subequations}\label{equ: mmp formulation}
    \begin{align}
    \bm{\tau}^* &= \argmax_{\bm{\tau}}\ \log p(\bm{\tau}|\mathcal{O}), \\
    &= \argmax_{\bm{\tau}}\log p(\mathcal{O}|\bm{\tau})+\log p(\bm{\tau}),\label{equ: map factorized} \\
    \mathrm{s.t.} & \quad \  \bm{\tau}\in\mathcal{F} , \label{equ: map feas set}
\end{align}
\end{subequations}
where $p(\mathcal{O}|\bm{\tau})$ is the trajectory cost, $p(\bm{\tau})$ is a prior of trajectories that can be learned from real-world or simulated data, and $\mathcal{F}$ is a set of hard constraints, e.g., terminal or formation. Thus, in the hybrid optimization-generative approach, generative models provide informative priors that accelerate the convergence towards high-likelihood regions. An alternative view is to guide the sampling of the generative model to meet the task-specific objectives \cite{carvalho2023motion} 

\subsection{Decoupling Multiagent Objectives}

Since it is difficult to directly learn a joint prior $p(\bm{\tau})=p(\bm{\tau}_1,\dots,\bm{\tau}_N)$ for multiple agents, we decompose the task such that single-agent priors are compatible. The set of objectives $\mathcal{O}$ admits
\begin{equation}\label{equ: int and intr obj}
    \mathcal{O} = \bigcup_{i=1}^{N} \mathcal{O}_i,
\end{equation}
where $\mathcal{O}_i = \qty{o_{i,1},\dots,o_{i,K_i}}$ contains $K_i$ individual objectives for agent $i$. We then make the following assumption on the decomposition of $\mathcal{O}$.
\begin{assumptionx}[\textit{Conditional independence of objectives and local dependence of trajectories}]\label{asmp: decoupling obj}
Given the joint trajectory $\boldsymbol{\tau}$, all task objectives are conditionally independent; for example, trajectory smoothness does not provide information about collision. Furthermore, objectives depend only on that agent's trajectory i.e., $p_{i,k}(o_{i,k}|\bm{\tau}) = p_{i,k}(o_{i,k}|\bm{\tau}_i)$.
\end{assumptionx}
\noindent Given Assumption \ref{asmp: decoupling obj}, the likelihood \eqref{equ: map factorized}  could be factorized as 
\begin{subequations}\label{equ: obj likelihood}
    \begin{align}
        p(\mathcal{O}|\bm{\tau}) &= p\qty(\qty{o_{i,k}}_{\substack{i=1:N \\ k=1:K_i}}\mid\bm{\tau}), \\
        &=\prod_{i=1}^N\prod_{k=1}^{K_i}p_{i,k}(o_{i,k}\mid\bm{\tau}_i).
\end{align}
\end{subequations}

\begin{assumptionx}[\textit{Exponential form of objectives \cite{carvalho2023motion}}]\label{asmp: exp obj}
We assume the likelihood of satisfying objectives $o_{i,k}$ for $i=1,\dots,N$, $k=1,\dots,K_i$ takes the exponential form $p_{i,k}(o_{i,k}|\bm{\tau}_i) \propto \exp\qty(-\lambda_{i,k}c_{i,k}(\bm{\tau}_i))$, where $c_{i,k}$ is the corresponding cost function with $\lambda_{i,k} > 0$ being the respective tuning weights.
\end{assumptionx}
\noindent This common assumption bridges the deterministic cost functions of the system with their statistical form. Then, the log-likelihood form of (\ref{equ: obj likelihood}) can be further simplified to \begin{equation}\label{equ: multiagent cost}
    \log p(\mathcal{O}\mid\bm{\tau}) \propto - \sum_{i=1}^N\sum_{k=1}^{K_i}\lambda_{i,k}c_{i,k}(\bm{\tau}_i),
\end{equation}
where the per-agent costs $c_{i,k}(\bm{\tau}_i)$ include obstacle collision cost using the signed distance field (SDF) and a Gaussian process (GP) smoothness cost, which we adopt from \cite{carvalho2023motion}.

Since we have decomposed the high-dimensional multiagent objectives, we can now apply the single-agent hybrid planning workflow to the decoupled objectives. Next, we orient through the single-agent planning workflow with diffusion models.

\subsection{Single-agent Motion Planning Diffusion}
\begin{figure}[t]
    \centering%
    \includegraphics[width=1\linewidth]{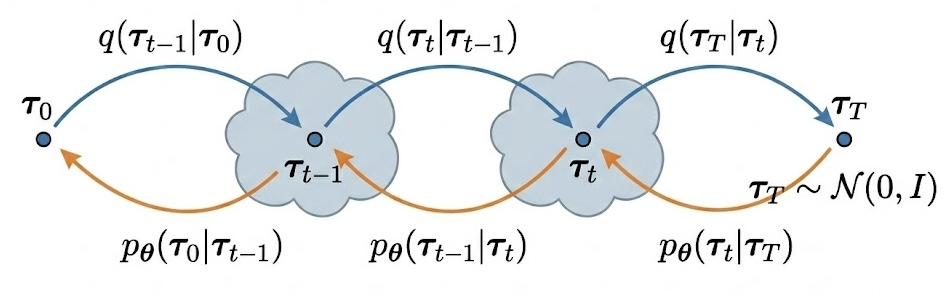}
    \caption{Forward and reverse diffusion processes: noise injection (blue arrows) and iterative denoising (orange arrows)}%
    \label{fig: diagram diffusion}%
\end{figure}

In this section, we sketch the essentials of single-agent motion planning diffusion without the intricate mechanics. Readers may refer to \cite{ho2020denoising, carvalho2023motion} for additional details. An illustration of the process is shown in Fig. \ref{fig: diagram diffusion}.

To learn the single-agent prior $p(\bm{\tau}_i)$ from data, a diffusion model progressively adds Gaussian noise to a clean trajectory $\bm{\tau}_{i,0}$ through a $T$-step Markovian forward process. Let $\bm{\tau}_{i,t}$ denote the trajectory of agent $i$ at diffusion step $t=1,\dots,T$. The forward transition is defined as 
\begin{equation}
    q(\bm{\tau}_{i,t}|\bm{\tau}_{i,t-1}) = \mathcal{N}\qty(\bm{\tau}_{i,t}; \sqrt{\alpha_t}\bm{\tau}_{i,t-1}, (1-\alpha_t)\bm{I}),
\end{equation}
where $\alpha_t$ is a predefined noise schedule. This formulation allows us to derive $\bm{\tau}_{i,t}$ directly from the clean data $\bm{\tau}_{i,0}$ via
\begin{equation}\label{equ: forward diffusion direct}
    \bm{\tau}_{i,t} = \sqrt{\bar{\alpha}_t}\bm{\tau}_{i,0} + \sqrt{1-\bar{\alpha}_t}\bm{\epsilon}_0 \sim \mathcal{N}\qty(\bm{\tau}_{i,t}; \sqrt{\bar{\alpha}_t}\bm{\tau}_{i,0}, (1-\bar{\alpha}_t)\bm{I}),
\end{equation}
where $\bar{\alpha}_t = \prod_{s=1}^t \alpha_s$ and the source noise is $\bm{\epsilon}_0 \sim \mathcal{N}(\bm{0}, \bm{I})$. 

Conversely, the generative reverse process recovers the clean trajectory from standard Gaussian noise $\bm{\tau}_{i,T} \sim \mathcal{N}(\bm{0},\bm{I})$ through a sequence of denoising transitions $p_{\bm{\theta}}(\bm{\tau}_{i,t-1}|\bm{\tau}_{i,t})$, which We model as Gaussian, parameterized by a neural network with weights $\bm{\theta}$ i.e.,
\begin{equation}\label{equ: Gauss denoise dist}
    p_{\bm{\theta}}(\bm{\tau}_{i,t-1}|\bm{\tau}_{i,t}) = \mathcal{N}\qty(\bm{\tau}_{i,t-1}; \bm{\mu}_{\bm{\theta}}(\bm{\tau}_{i,t}, t), \bm{\Sigma}_t).
\end{equation}
The covariance is typically fixed to $\bm{\Sigma}_t = \sigma_t^2\bm{I}$, where $\sigma_t^2 = \frac{(1-\alpha_t)(1-\bar{\alpha}_{t-1})}{1-\bar{\alpha}_t}$. The mean of the reverse process is predicted by a noise-estimation network $\hat{\bm{\epsilon}}_\theta(\bm{\tau}_{i,t}, t)$, given by
\begin{equation}
    \bm{\mu}_{\bm{\theta}}(\bm{\tau}_{i,t}, t) = \frac{1}{\sqrt{\alpha_t}}\bm{\tau}_{i,t} - \frac{1-\alpha_t}{\sqrt{1-\bar{\alpha}_t}\sqrt{\alpha_t}}\hat{\bm{\epsilon}}_\theta(\bm{\tau}_{i,t}, t).
\end{equation}
The network parameters $\bm{\theta}$ are trained using a loss function that minimizes the difference between the true source noise and the predicted noise:
\begin{equation}
    \mathcal{L}(\bm{\theta}) = \mathbb{E}_{t,\bm{\epsilon}_0,\bm{\tau}_{i,0}}\norm{\hat{\bm{\epsilon}}_{\bm{\theta}}(\bm{\tau}_{i,t}, t) - \bm{\epsilon}_0}_2^2,
\end{equation}
where $t$ is sampled uniformly from $\{1, \dots, T\}$.

While the trained network captures the trajectory prior $p(\bm{\tau}_i)$, single-agent planning requires conditioning this generation on the local task objectives $\mathcal{O}_i$. Applying Bayes' rule at each denoising step yields the task-conditioned posterior 
\begin{equation}\label{equ: task conditioned posterior}
    p_{\bm{\theta}}(\bm{\tau}_{i,t-1}|\bm{\tau}_{i,t}, \mathcal{O}_i) \propto p_{\bm{\theta}}(\bm{\tau}_{i,t-1}|\bm{\tau}_{i,t}) p(\mathcal{O}_i|\bm{\tau}_{i,t-1}).
\end{equation}
Now, to evaluate the likelihood $p(\mathcal{O}_i|\bm{\tau}_{i,t-1})$, we approximate the task log-likelihood with a first-order Taylor expansion around the unconditional mean $\bm{\mu}_{i,t} = \bm{\mu}_{\bm{\theta}}(\bm{\tau}_{i,t}, t)$, which gives
\begin{equation}
    \log p(\mathcal{O}_i|\bm{\tau}_{i,t-1}) \approx \log p(\mathcal{O}_i|\bm{\mu}_{i,t}) + (\bm{\tau}_{i,t-1} - \bm{\mu}_{i,t})^\top \bm{g}_i,
\end{equation}
where the gradient $\bm{g}_i = \nabla_{\bm{\tau}_{i,t-1}} \log p(\mathcal{O}_i|\bm{\tau}_{i,t-1})\big|_{\bm{\tau}_{i,t-1}=\bm{\mu}_{i,t}}$ incorporates the decoupled intra-agent costs
\begin{equation}\label{equ: single agent guidance gradient}
    \bm{g}_i = -\sum_{k=1}^{K_i}\lambda_{i,k}\nabla_{\bm{\tau}_i} c_{i,k}(\bm{\mu}_{i,t}).
\end{equation}
Combining this linear approximation with the Gaussian transition in (\ref{equ: Gauss denoise dist}) results in a shifted Gaussian distribution for the guided denoising step:
\begin{equation}
    p_{\bm{\theta}}(\bm{\tau}_{i,t-1}|\bm{\tau}_{i,t}, \mathcal{O}_i) \approx \mathcal{N}\qty(\bm{\tau}_{i,t-1}; \bm{\mu}_{i,t} + \bm{\Sigma}_t \bm{g}_i, \bm{\Sigma}_t).
\end{equation}
Consequently, we can generate task-satisfying trajectories for a single agent by iteratively sampling from this distribution. As can be seen, the mean $\bm{\mu}_{i,t} + \bm{\Sigma}_t \bm{g}_i$ combines the diffusion denoising and the gradient of the objectives, which is the backbone of the hybrid optimization-generative approach. 

The single-agent diffusion process only tackles the intra-agent objective, such as collision avoidance with obstacles, smoothness, etc. Inter-agent interactions, such as the formation and collision avoidance are formulated as constraints and will be introduced in the next section.

\section{Formational Motion Planning Diffusion}\label{sec: fmpd}
This section generalizes the single-agent motion planning diffusion to the multiagent case, particularly under the formation constraints. More specifically, the diffusion priors still fulfill the individual objectives, while the agents' interactions are handled through formation constraints $\mathcal{F}$ in (\ref{equ: map feas set}). By interleaving the guided reverse diffusion with projected gradient descent (PGD), we systematically project the multiagent trajectories onto the feasible formation manifold. Note that the diffusion with PGD has been initially attempted in \cite{Liang2025SMD} however, it has not been explored for FMP problems.

\subsection{Joint Posterior for Multiagent Planning}
Let $\bm{\tau}_t \in \mathbb{R}^{2DHN}$ denote the joint sample at diffusion step $t=1,\dots,T$, which concatenates the ensemble of trajectories for all $N$ agents. Similar to (\ref{equ: task conditioned posterior}), the task-conditioned posterior of the joint denoising step follows Bayes' rule:
\begin{equation}
    p_{\bm{\theta}}(\bm{\tau}_{t-1}|\bm{\tau}_t, \mathcal{O}) \propto p_{\bm{\theta}}(\bm{\tau}_{t-1}|\bm{\tau}_t) p(\mathcal{O}|\bm{\tau}_{t-1}).
\end{equation}
Since the unconditioned trajectory generation is independent across agents, the joint unconditioned denoising transition $p_{\bm{\theta}}(\bm{\tau}_{t-1}|\bm{\tau}_t)$ remains Gaussian, given by
\begin{equation}
    \log p_{\bm{\theta}}(\bm{\tau}_{t-1}|\bm{\tau}_t) \propto -\frac{1}{2}(\bm{\tau}_{t-1} - \bm{\mu}_t)\tp \bm{\Sigma}_t^{-1} (\bm{\tau}_{t-1} - \bm{\mu}_t),
\end{equation}
where the joint mean and covariance take the block-structured forms
\begin{subequations}
    \begin{align}
        \bm{\mu}_t &= [\bm{\mu}_{1,t}^\top, \dots, \bm{\mu}_{N,t}^\top]^\top \in \mathbb{R}^{2DHN}, \label{equ: joint mean} \\
        \bm{\Sigma}_t &= \mathbf{I}_N \otimes \sigma_t^2 \mathbf{I}_{2DH}. \label{equ: joint cov}
    \end{align}
\end{subequations}
Here, $\bm{\mu}_{i,t} = \bm{\mu}_{\bm{\theta}}(\bm{\tau}_{i,t}, t)$ represents the denoising mean for agent $i=1,\dots,N$. Note that each agent applies the same denoiser trained from the single-agent dataset. 

Again, the task likelihood is approximated using a first-order Taylor expansion around $\bm{\mu}_t$, the task likelihood is approximated as $\log p(\mathcal{O}|\bm{\tau}_{t-1}) \approx \log p(\mathcal{O}|\bm{\mu}_t) + (\bm{\tau}_{t-1} - \bm{\mu}_t)^\top \bm{g}$, where the joint guidance gradient is $\bm{g} = \nabla_{\bm{\tau}_{t-1}} \log p(\mathcal{O}|\bm{\tau}_{t-1})\big|_{\bm{\tau}_{t-1}=\bm{\mu}_t}$. Combining this with the prior yields a shifted joint Gaussian:
\begin{equation}
    p_{\bm{\theta}}(\bm{\tau}_{t-1}|\bm{\tau}_t, \mathcal{O}) \propto \mathcal{N}\qty(\bm{\tau}_{t-1}; \bm{\mu}_t + \bm{\Sigma}_t \bm{g}, \bm{\Sigma}_t).
\end{equation}
Since the covariance $\bm{\Sigma}_t$ is block-diagonal, the mean shift decouples strictly across the agents, i.e., $\bm{\Sigma}_t \bm{g} = [\sigma_t^2 \bm{g}_1\tp, \dots, \sigma_t^2 \bm{g}_N\tp]\tp$. This structural property splits the joint Gaussian posterior into a product of independent, parallel distributions:
\begin{equation}
    p_{\bm{\theta}}(\bm{\tau}_{t-1}|\bm{\tau}_t, \mathcal{O}) = \prod_{i=1}^N \mathcal{N}\qty(\bm{\tau}_{i,t-1}; \bm{\mu}_{i,t} + \sigma_t^2 \bm{g}_i, \sigma_t^2 \mathbf{I}_{2DH}).
\end{equation}
As derived from the factorized multiagent likelihood in (\ref{equ: multiagent cost}), the gradient per-agent $\bm{g}_i$ admits
\begin{equation}\label{equ: multiagent guidance gradient}
    \bm{g}_i = -\sum_{k=1}^{K_i}\lambda_{i,k}\nabla_{\bm{\tau}_i} c_{i,k}(\bm{\mu}_{i,t}).
\end{equation}

\subsection{Affine Formation Constraints with Safety Guarantees}\label{sec: projection with safety}
While the soft costs in (\ref{equ: multiagent guidance gradient}) effectively manage the individual objectives, we formulate the formation requirement as hard constraints. Recollect that the state $\bm{s}_{i,h} = \mqty[\bm{p}_{i,h}\tp & \dot{\bm{p}}_{i,h}\tp]\tp\in\mathbb{R}^{2D}$ contains the position $\bm{p}_{i,h}\in\mathbb{R}^D$ of agent $i=1,\dots,N$ at waypoint $h=1,\dots,H$. We stack the positions of all $N$ agents column-wise into the \textit{configuration matrix},
\begin{equation}
    \bm{P}_h = \qty[\bm{p}_{1,h}, \dots, \bm{p}_{N,h}] \in \mathbb{R}^{D \times N},
\end{equation}
which should be a continuous transformation of a template, namely the \textit{nominal configuration}. Typically, the nominal configuration is chosen as the initial configuration $\bm{P}_0 = \qty[\bm{p}_{1,0}, \dots, \bm{p}_{N,0}]\in\mathbb{R}^{D\times N}$. Then the transformation could be described by a general affine transformation
\begin{equation}\label{equ: general affine trans}
    \bm{P}_h = \bm{A}_h \bm{P}_0 + \bm{b}_h \bm{1}_N\tp,
\end{equation}
where $\bm{A}_h\in\mathbb{R}^{D\times D}$ and $\bm{b}_h\in\mathbb{R}^D$ are the shape transformation and the translations, respectively. Now, we seek a projection operator $\Pi_{\mathcal{F}}(\bm{P}_h)$ that satisfies (\ref{equ: general affine trans}).

We now define the augmented nominal configuration \begin{equation}
    \bm{Q} = \mqty[\bm{P}_0\tp & \bm{1}_N]\tp \in \mathbb{R}^{(D+1) \times N},
    \label{equ:Q_definition}
\end{equation} and a compact notation $\bm{\Theta}_h = \mqty[\bm{A}_h & \bm{b}_h]\in\mathbb{R}^{D\times(D+1)}$. Then the affine constraint \eqref{equ: general affine trans} can be written compactly as $\bm{P}_h = \bm{\Theta}_h\bm{Q}$. A naive approach would project each waypoint independently by solving $\bm{\Theta}_h = \bm{P}_h\bm{Q}\tp(\bm{Q}\bm{Q}\tp)^{-1}$, which ignores the temporal coherence, potentially resulting in erratic behaviors in the trajectories. To overcome this limitation, we propose a temporally-regularized fit as follows
\begin{equation}\label{equ: regularized fit}
\min_{\qty{\bm{\Theta}_h}_{h=0}^{H-1}} \sum_{h=0}^{H-1} \norm{\bm{P}_h - \bm{\Theta}_h \bm{Q}}\Fro^2 + \eta r\qty(\qty{\bm{\Theta}_h}_{h=0}^{H-1})
\end{equation}
where the regularization term $r(\cdot)$ can be any smoothing function over a discrete set of affine parameters $\bm{\Theta}_h$ for $h=0,\dots,H-1$, weighted by a hyperparameter $\eta$. In practice, $\bm{\Theta}_h$ could also be separated into $\bm{A}_h$ and $\bm{b}_h$, and estimated individually whose weights are tuned independently.

The advantage of navigating a swarm of agents in formation is that the agents' separation is directly related to the geometry setting $\bm{A}_h$, which we could leverage to guarantee inter-agent collision avoidance. Observe that 
\begin{subequations}\label{equ: spacing bound}
    \begin{align}
    \norm{\bm{p}_{i,h}-\bm{p}_{j,h}} &= \norm{\bm{A}_h(\bm{p}_{i,0}-\bm{p}_{j,0})},\\
    &\geq \sigma_{\min}(\bm{A}_h)\norm{\bm{p}_{i,0}-\bm{p}_{j,0}}, \\ &\geq \sigma_{\min}(\bm{A}_h)\,d_0,
\end{align}
\end{subequations}
where $\sigma_{\min}(\bm{A}_h)$ is the smallest singular value of $\bm{A}_h$, and $d_0 = \min_{i\neq j}\norm{\bm{p}_{i,0}-\bm{p}_{j,0}}_2$ is the minimum separation in the nominal template. Thus, by flooring the smallest singular value of $\bm{A}_h$ from the projected configuration, we could ensure a minimum safety distance $d_{\mathrm{safe}}$. As such, given the generated configuration $\bm{P}_h$ and a initial projection $\bm{\Theta}_h = \mqty[\bm{A}_h & \bm{b}_h]$ from \eqref{equ: regularized fit}, we could further perform a clamping
\begin{subequations}\label{equ: singular value clamp}
    \begin{align}
        \bm{A}_h &= \bm{U}\bm{\Sigma}\bm{V}\tp,\ \bm{\Sigma} = \diag(\sigma_1,\dots,\sigma_D), \\
        \underline{\bm{\Sigma}_h} &= \diag\qty(\max(\sigma_1,d_{\mathrm{th}}),\dots,\max(\sigma_D,d_{\mathrm{th}})),\\
        \underline{\bm{A}_h} &= \bm{U} \underline{\bm{\Sigma}_h}\bm{V}\tp,
    \end{align}
\end{subequations}
where $d_{\mathrm{th}} = d_{\mathrm{safe}}/d_0$. Finally,
the smooth affine formation projection with safety guarantees can be written as 
\begin{equation}\label{equ: formation proj}
    \Pi_{\mathcal{F}}(\bm{P}_h) = \underline{\bm{A}_h}\bm{P}_0 + \bm{b}_h\bm{1}_N\tp.
\end{equation}
Note that in practice, the singular value clamping might also disrupt the trajectory smoothness, which could be effectively mitigated with a few iterations of alternating \eqref{equ: singular value clamp} and \eqref{equ: regularized fit}.

The data generation and the training parts of the diffusion model match the setup from Algorithm 1 in \cite{carvalho2023motion}, and are thus not discussed in detail. Algorithm \ref{alg: FMPD} outlines the inference part of the framework, which generally follows a sequence of diffusion steps, gradients, and projections as discussed previously. The flexibility in the implementation lies in applying the gradient steps, i.e., we could apply multiple gradient steps after one diffusion step, while applying no gradient steps for the first few diffusion steps. This flexibility shows better computational efficiency and performance in existing literature.
\begin{algorithm}[t]
    \caption{\textit{Formational Motion Planning Diffusion}}\label{alg: FMPD}
    \begin{algorithmic}[1]
        \State \textbf{Input}: Pre-trained diffusion model $\hat{\bm{\epsilon}}_{\bm{\theta}}$; Nominal configuration template $\bm{P}_0$ and the goal configuration $\bm{P}_H$; Intra-agent costs $c_{i,k}$ with weights $\lambda_{i,k}$; Inter-agent costs $c_{ij}$ with weights $\lambda_{ij}$; scheduling terms $\alpha_t, \sigma_t^2$
        \State $\rhd$ \textit{Sample initial joint noise}
        \State $\bm{\tau}_{i,T} \sim \mathcal{N}(\bm{0}_{2DH}, \bm{I}) \;\; \forall i = 1,\ldots,N$
        \State $\rhd$ \textit{Enforce boundary conditions}
        \State $\bm{\tau}_T[i, 0] = \bm{p}_{i,0},\quad \bm{\tau}_T[i, H-1] = \bm{p}_{i,H-1} \;\; \forall i \le N$
        \For{$t = T, \ldots, 1$}
            \State $\rhd$ \textit{Diffusion prior mean}
            \State $\bm{\mu}_{i,t} = \dfrac{1}{\sqrt{\alpha_t}}\qty(\bm{\tau}_{i,t} - \dfrac{1 - \alpha_t}{\sqrt{1 - \bar{\alpha}_t}}\hat{\bm{\epsilon}}_{\bm{\theta}}(\bm{\tau}_{i,t}, t)) \;\; \forall i \le N$
            \State Compute $\bm{g}_i\ \forall i \le N $ based on \eqref{equ: multiagent guidance gradient}
            \State $\rhd$ \textit{Unconstrained update}
            \State $\widetilde{\bm{\tau}}_{i,t-1} = \bm{\mu}_{i,t} + \sigma_t^2\,\bm{g}_i + \sigma_t \bm{\varepsilon}_i, \; \bm{\varepsilon}_i \sim \mathcal{N}(\bm{0}, \bm{I}) \;\; \forall i \le N$
            \State $\rhd$ \textit{Formation Projection}
            \For{waypoint $h = 1, \dots, H-2$}
                \State $\bm{P}_h = [\bm{p}_{1,h} \cdots \bm{p}_{N,h}]$
                \State Projection using \eqref{equ: regularized fit} and \eqref{equ: formation proj} $\bm{P}_h \leftarrow \Pi_{\mathcal{F}}(\bm{P}_h)$
                \State Reassemble projected states back into $\widetilde{\bm{\tau}}_{t-1}$
            \EndFor
            
            \State $\rhd$ \textit{Update trajectory}
            \State $\bm{\tau}_{i,t-1} = \widetilde{\bm{\tau}}_{i,t-1} \;\; \forall i \le N$
        \EndFor

        \State \textbf{Output}: Optimized formational plan $\bm{\tau}_0 = \{\bm{\tau}_{0,i}\}_{i=1}^{N}$
    \end{algorithmic}
\end{algorithm}

\section{Simulations}\label{sec: sim}
In this section, we demonstrate our planner using a $7$-agent formation in a confined environment, and we also provide the results for two other state-of-the-art methods, Du et al. \cite{du2025motion} and Quan et al. \cite{quan2023robust}, representing the two classic paradigms of complexity reduction discussed in Section \ref{sec: intro}.  Apart from the planner results, we also show that the planned trajectories are easy to track by a simple proportion-derivative (PD)-based distributed controller under double-integrator dynamics \cite{zhao2018affine}.

\subsection{Experimental Setup}
The environment is a square of $20m$ and the agents are 2D disks with a radius of $0.2$m. We then set the minimum safety distance $d_{\text{safe}}=0.45m$, and all agents are color-coded in the plots. Our proposed planner is given a horizon of $H=128$ with a duration of $30$s, resulting in a discrete-time interval $\Delta t = 0.23s$. We run $T = 25$ denoising steps for the diffusion model. The results are run on a workstation equipped with an AMD Ryzen 7900X CPU with $64$ GB RAM and an NVIDIA RTX PRO 4500 Blackwell GPU with $32$ GB of VRAM. 

For the smoothing regularization function proposed in \eqref{equ: regularized fit}, we opt for a discrete second-order derivative  $r(\qty{\bm{\Theta}_h}_{h=0}^{H-1}) = \sum_{h=1}^{H-2} \norm{\bm{\Theta}_{h+1} - 2\bm{\Theta}_h + \bm{\Theta}_{h-1}}\Fro^2$, which promotes smooth geometry transition across the horizon. To execute the planned trajectories, we employ a discrete-time version \cite{xu2020affine} of the classic distributed affine formation controller \cite{zhao2018affine} under discrete double-integrator dynamics $\bm{z}_{i,h+1} = \bm{z}_{i,h} +\Delta t\bm{v}_{i,h},\bm{v}_{i,h+1} = \bm{v}_{i,h} + \Delta t\bm{u}_{i,h}$, where $\bm{z}_{i,h}\in\mathbb{R}^2$ is the real position of agent $i$ at waypoint $h$. Each leader $i$, then tracks its reference trajectory $\bm{p}_{i,h}$ under the control law
\begin{equation}\label{equ: leader control}
   \bm{u}_{i,h} = -k_p\qty(\bm{z}_{i,h} - \bm{p}_{i,h}) - k_v(\bm{v}_{i,h} - \dot{\bm{p}}_{i,h}) + \dot{\bm{p}}_{i,h}, 
\end{equation} where $k_p$ and $k_v$ are the tuning gains, and $\bm{p}_{i,h}$ with its derivative $\dot{\bm{p}}_{i,h}$ are the planned states. For a follower $i$, which interacts with a set of neighboring agents $\mathcal{N}_i$ predefined by the underlying communication graph, the corresponding control law is 
\begin{align}\label{equ: follower control}
    \bm{u}_{i,h} =& -\frac{1}{\gamma_i}\sum_{j\in\mathcal{N}_i}\omega_{ij}(k_p(\bm{z}_{i,h} - \bm{z}_{j,h}) \notag \\
    & + k_v(\bm{v}_{i,h} - \bm{v}_{j,h}) - \dot{\bm{v}}_{j,h}),
\end{align}
where $\gamma_i = \sum_{j\in\mathcal{N}_i}\omega_{ij}$, and $\omega_{ij}$ are the equilibrium stresses which can be computed using e.g., framework \cite{li2025fast}.

\begin{figure*}[t]
    \centering%
    \includegraphics[width=1\linewidth]{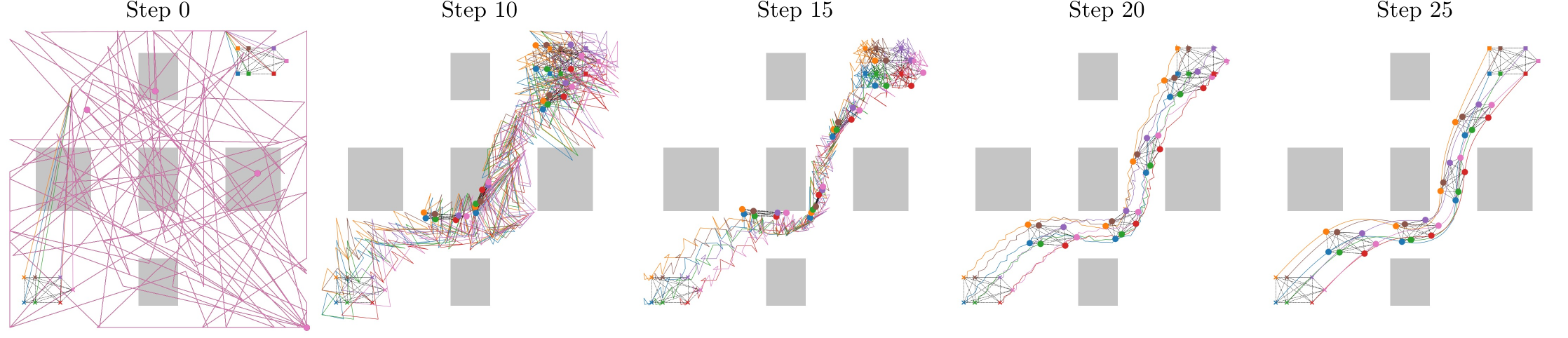}
    \caption{The multiagent trajectories (color-coded) at different diffusion inference steps, with some intermediate formations outlined with dashed lines. The gray areas are the obstacles.}%
    \label{fig: diffusion steps}%
\end{figure*}

\subsection{Evaluation Metrics}
We propose the following metrics to quantify the quality of the generated trajectories.

\noindent \textbf{Affine Residual (AR)} measures the instantaneous geometry of the trajectories, aiming to evaluate how much the formation is compromised in resolving collisions. It is computed by measuring the distance of the configuration from the affine set using the orthogonal projector $\bm{M} = \bm{I}_N - \bm{Q}^\top(\bm{Q}\bm{Q}^\top)^{-1}\bm{Q}$, where $\bm{Q}\in\mathbb{R}^{(D+1)\times N}$ is the augmented nominal configuration (\ref{equ:Q_definition}). Then the error $\delta_{\text{ar}}$ is defined as
\begin{equation}
    \delta_{\text{ar}} = \frac{1}{NH}\sum_{h=0}^{H-1}\norm{\bm{P}_h\bm{M}}^2\Fro,
\end{equation}
which equals zero if every waypoint across the horizon  is an exact affine transformation of the nominal template. Note that this metric also applies to the compared candidates \cite{du2025motion,quan2023robust}, since their formation geometries are special cases of the general affine transformation with reduced degrees of freedom.

\noindent \textbf{Smoothness (S)} in $[m/s^2]$ evaluates the temporal smoothness of the generated trajectories by leveraging the discrete second difference of the positions for each agent, denoted as $\Delta^2 \bm{p}_{i,h} = \bm{p}_{i,h+1} - 2\bm{p}_{i,h} + \bm{p}_{i,h-1}$. Then the smoothness metric is expressed as 
\begin{equation}
    \delta_{\text{s}} = \sqrt{\frac{1}{N(H-2)}\sum_{i=1}^{N}\sum_{h=1}^{H-2} \norm{ \frac{\Delta^2 \mathbf{p}_{i,h}}{\Delta t^2} }_2^2},
\end{equation}
where $\Delta t$ is the discrete time interval. Trajectories are considered smoother for lower values of $\delta_{\text{s}}$.

\noindent \textbf{Tracking Error (TE)} in $[m]$ evaluates how well the formation maneuvers can be tracked using basic distributed controllers \eqref{equ: leader control} and \eqref{equ: follower control}. For individual agents, TE is defined as ${\delta_{te}}({i,h}) = \norm{\bm{z}_{i,h} - \bm{p}_{i,h}}_2$.

\noindent \textbf{Collision (C)} in [\%] evaluates the percentage of waypoints that violate the minimum clearance distance given the robot model and the obstacles.

\noindent \textbf{Planning Time (PT)} in $[s]$ is the computation time for the planning. This metric provides a general insight into the complexity of the algorithms. We do not draw any comparative conclusions with the state-of-the-art, as \cite{du2025motion} is an online planner and admits distributed executions, and we translate \cite{quan2023robust}'s own implementation from C++ to Python for our test environment.

\subsection{Results and Discussions}
We first show the process of the proposed diffusion-based planner in Fig. \ref{fig: diffusion steps}, where the trajectories are gradually denoised from random Gaussian initialization to clean samples. Observe that at initial steps, e.g., Step 10, there are still collisions with the obstacles, but the trajectories are nudged away from the obstacles as a mutual effort of the diffusion priors and the cost gradients.

\begin{figure}[hbt]
    \centering%
    \includegraphics[width=1\linewidth]{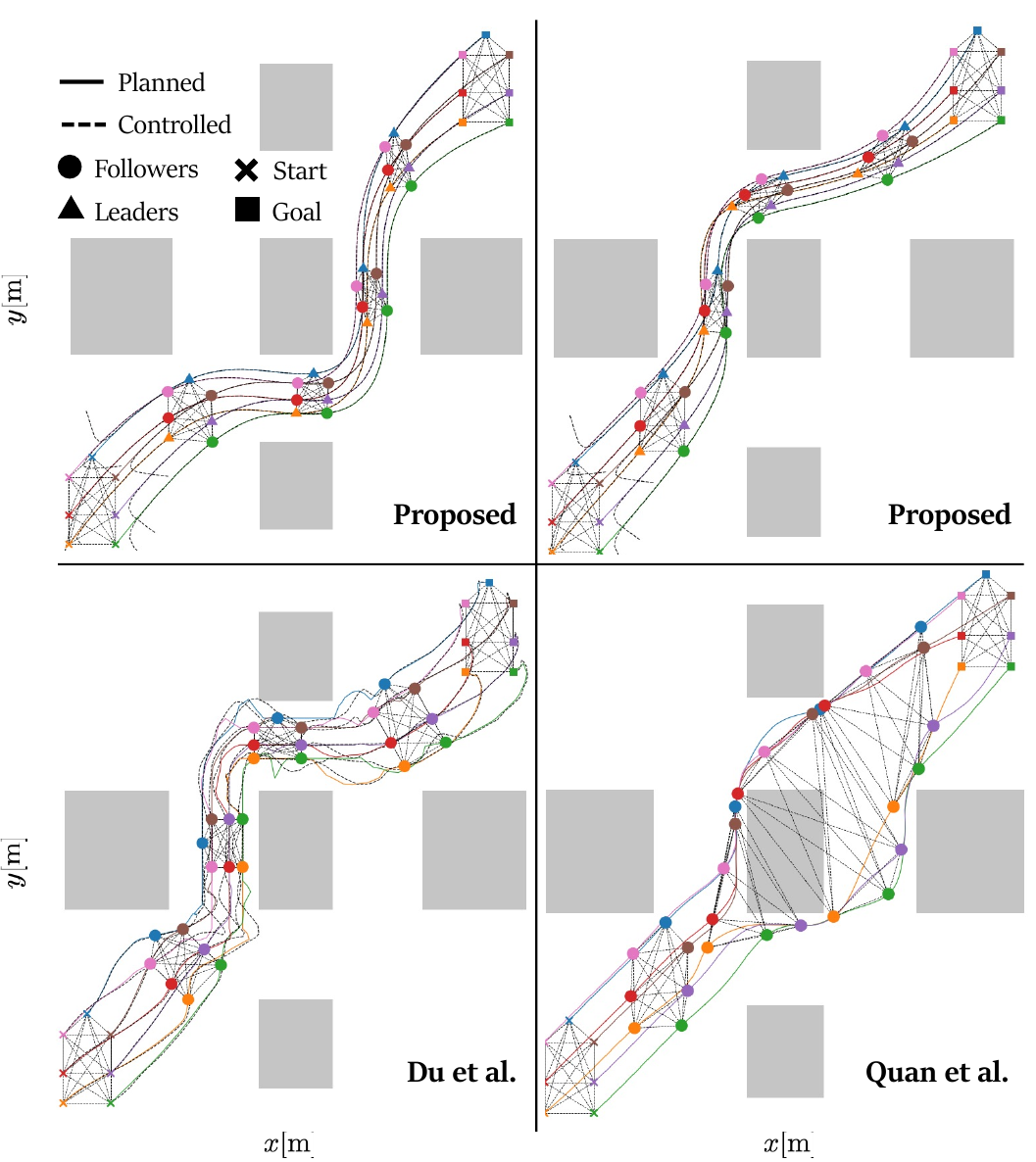}
    \caption{The planned and controlled trajectories of the proposed (upper) and state-of-the-art (lower). The proposed planner is tracked with a basic controller \eqref{equ: leader control} and \eqref{equ: follower control}; Du et al. \cite{du2025motion} is tracked with its own MPC; Quan et al. \cite{quan2023robust} is tracked using designated quadrotor controllers, hence not implemented.}
    \label{fig: fmpd with control}%
\end{figure}

\begin{figure}[t]
    \centering%
    \includegraphics[width=1\linewidth]{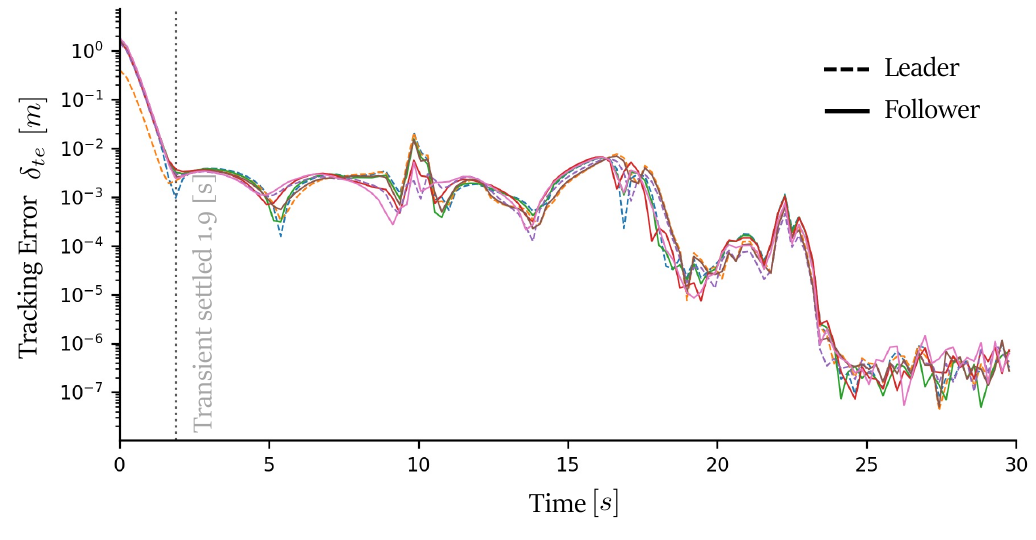}
    \caption{Agents' tracking error over time for the trajectories using the proposed planner under tracking controllers \eqref{equ: leader control} and \eqref{equ: follower control}.}%
    \label{fig: fmpd tracking error}%
\end{figure}

Next, we show a few more examples of the proposed solutions and the state-of-the-art for both the planned and controller-tracked trajectories in Fig. \ref{fig: fmpd with control}. The proposed solution does not lead to deterministic results, as can be seen in the top row of Fig. \ref{fig: fmpd with control}, where the swarm could either pass from the left or right corridor given different randomness. This multi-modality is favored in motion planning for reachability and avoiding local minima. From a trajectory quality perspective, the proposed planner and \cite{quan2023robust} yield visibly smoother and more natural trajectories compared to \cite{du2025motion}, which relies on a coarse artificial potential field (APF) planner. From a formation constraints perspective, the proposed method admits rigorous affine transformation of the nominal template, and \cite{du2025motion} can also maintain its desired geometry, i.e., rotation with anisotropic scaling. However, \cite{quan2023robust} trades off the formation geometry for maneuverability across the passages, as formation is never strictly enforced in their framework. From a tracking control perspective, the planned trajectories overlap with the controlled trajectories after initial convergence from random positions, which shows trackability for the proposed planner. The tracking error $\delta_{te}$ for each agent can be seen in Fig. \ref{fig: fmpd tracking error}, where the agents rapidly converge to their target positions and maintain a low tracking error ($\leq 0.02m$). As a comparison, \cite{du2025motion}'s MPC tracker partially corrects the coarse planning, hence the misalignment between the planned and the controlled trajectories in Fig. \ref{fig: fmpd with control}.

Lastly, we present the quantitative evaluations in Table ~\ref{tab: comparison dual passage}, where all results are the mean of $50$ Monte Carlo runs given different randomness. As indicated, the proposed planner has the best affine residual and smoothness among all. This shows the effectiveness of the generative prior and its compatibility with external objectives, especially in use of aggressive projections.  Moreover, the safe formation projection guarantees collision-free interactions, while the framework also keeps collisions with the obstacles to a minimum. Finally, recall that the joint trajectory $\bm{\tau}\in\mathbb{R}^{2DHN = 3584}$ is high-dimensional and usually intractable to directly optimize. Here, our proposed solution takes less than a second to plan for a $7$-agent formation, and we also recorded less than $1.5s$ on average to plan for $50$ agents. The most computational complexity is offloaded to the training phase, such that the inference speed is primarily bottlenecked by the size of the neural network $\bm{\theta}$. Compared with \cite{du2025motion}, where the tracking requires significantly more effort than the planning, we could claim that high-quality planning can offload the complexity on the control end, which consumes less onboard resource for these autonomous agents. The low tracking error also verifies this claim.

\begin{table}[t]
\centering
\caption{Quantitative planning performance}
\label{tab: comparison dual passage}
\setlength{\tabcolsep}{5pt}
\begin{tabular}{lccccc}
\toprule
Method & AR & S $[m/s^2]$ & TE $[m]$ & C $[\%]$ & PT $[s]$ \\
\midrule
Proposed & $\bm{3.0e{-13}}$ & $\bm{0.5}$ &  $\bm{1.5e{-3}}$ & ${0.08}$ & $\bm{0.97}$\\
Du et al. \cite{du2025motion}  & $3.7e{-3}$    & $1.9$      &   -               &  $0.2$ &$65.3^*$\\
Quan et al. \cite{quan2023robust} & $4.2e{-1}$  & $1.4$      & -                   & $\bm{0.00}$& $12.0$ \\
\bottomrule
\end{tabular}
\begin{tablenotes}
            \scriptsize
            \item[] $^*$ Actual PT occupies $0.2\%$ of the reported number; the rest is the MPC tracking. 
        \end{tablenotes}
\end{table}

\section{Conclusions}\label{sec: conclusion}
This work proposes a multiagent motion planning framework under affine formation constraints. The planning mixes projected gradient descent (PGD) and diffusion priors under a MAP formulation, where the diffusion prior is learned from the single-agent trajectory data and is shared across the agents for individual objectives, and PGD tackles the inter-agent interactions while maintaining feasible formations. Numerical results indicate our planner yields smooth and dynamically feasible trajectories under strict affine formation constraints with safety guarantees. The planner also admits moderate computation complexity which is tractable as the swarm size scale. Our future work includes a convergence analysis, physical simulation and experiments, and more comprehensive benchmarking.

\bibliographystyle{IEEEtran}
\bibliography{refs}

\end{document}